%% file: Submission.tex
\documentclass[10pt,twocolumn,letterpaper]{article}

\usepackage[pagenumbers]{cvpr} % To force page numbers, e.g. for an arXiv version

\usepackage{graphicx}
\usepackage{amsmath}
\usepackage{amssymb}
\usepackage{booktabs}

\usepackage{tabularx, booktabs, ragged2e}
\usepackage{adjustbox}
\usepackage{multirow}
\usepackage{lipsum}  
\usepackage[ruled,vlined]{algorithm2e}

\SetCommentSty{mycommfont}

\begin{document}

\title{Generating Topological Structure of Floorplans from Room Attributes}

\author{Yu Yin\\
Northeastern University, USA\\
{\tt\small yin.yu1@northeastern.edu}
\and
Will Hutchcroft,Naji Khosravan,Ivaylo Boyadzhiev\\
Zillow Group, USA\\
{\tt\small \{willhu,najik,ivaylob\}@zillowgroup.com}
\and
Yun Fu\\
Northeastern University, USA\\
{\tt\small yunfu@ece.neu.edu}
\and
~ ~ ~ ~ ~ ~ ~ ~ ~ ~ ~ ~ ~ ~ Sing Bing Kang~ ~ ~ ~ ~ ~ ~ ~ ~ ~ ~ ~ ~ ~ \\
Zillow Group, USA\\
{\tt\small @zillowgroup.com}
}

\maketitle

\begin{abstract}
    Analysis of indoor spaces requires topological information. 
    In this paper, we propose to extract topological information from room attributes using what we call Iterative and adaptive graph Topology Learning (ITL). ITL progressively predicts multiple relations between rooms; at each iteration, it improves node embeddings, which in turn facilitates generation of a better topological graph structure. This notion of iterative improvement of node embeddings and topological graph structure is in the same spirit as \cite{chen2020iterative}. However, while \cite{chen2020iterative} computes the adjacency matrix based on node similarity, we learn the graph metric using a relational decoder to extract room correlations. Experiments using a new challenging indoor dataset validate our proposed method. Qualitative and quantitative evaluation for layout topology prediction and floorplan generation applications also demonstrate the effectiveness of ITL.
\end{abstract}

%%%%%%%%% BODY TEXT
\input{sections/introduction}

\input{sections/relatedworks}

\input{sections/problem}

\input{sections/method}

\input{sections/experiments}
%-------------------------------------------------------------------------

\section{Conclusion}
In this paper, we propose to extract topological information from room attributes using what we call Iterative and adaptive graph Topology Learning (ITL). ITL progressively predicts multiple relations between rooms; at each iteration, it improves node embeddings, which allows for better topological structure learning and vise versa. In contrast to conventional similarity metric learning approaches (\eg \cite{chen2020iterative}), we learn the graph metric using a relational decoder to make better use of the containment relationship between the spatial and visual edges. 
% This results in more accurate room correlation predictions.
Experiments using a new challenging indoor dataset support our design decisions. Qualitative and quantitative evaluation for floorplan generation applications also demonstrates the effectiveness of ITL.

%%%%%%%%% REFERENCES
{\small
\bibliographystyle{ieee_fullname}
\bibliography{egbib}
}

% Supplementary Material
\clearpage
\section{Supplementary Material}
\appendix
% \chapter{Appendix Title}
\input{Supplementary}

\end{document}

%% file: sections/introduction.tex
\section{Introduction}
\label{sec:intro}
% Topological information of an architecture is an important prerequisite to the application of an indoor space.
%For a complex indoor environment, the proper modeling of each room as a separate sub-space reflects essential knowledge about its configuration. However, it is the most fundamental hierarchical correlation to decompose a top-down view of indoor space into independent rooms. Apart from room attributes, topological information of an architecture is another essential component to define and represent an indoor space, and it also plays a crucial role in designing, understanding, and remodeling the indoor environment. 

% Proper modeling of complex indoor environments is an essentials prerequisite for many applications (\eg indoor navigation~\cite{becker2009multilayered,jamali2017automated,mortari2019indoor}, house design~\cite{nauata2020house,hu2020graph2plan,para2020generative}, 3D reconstruction of indoor environments~\cite{tang2018full}). Encoding topological information can improve many of these applications while enabling other novel tasks. For example, inferring the most likely layout topology of an indoor space from a known set of rooms with their attributes, or completing a partially explored layout.

% Latest
Indoor home modeling enables practical applications such as virtual home tours, indoor navigation, and house design. Augmenting raw models with topological information enhances their value by providing semantic-level knowledge of room identity and arrangement. This, in turn, facilitates indoor understanding (e.g., style classification) and more home-related applications such as remote home shopping (through similarity-based search) and pricing.

\begin{figure}[t]
    \centering
    \includegraphics[trim=0in 0.2in 0in 0in,clip,width=.9\linewidth]{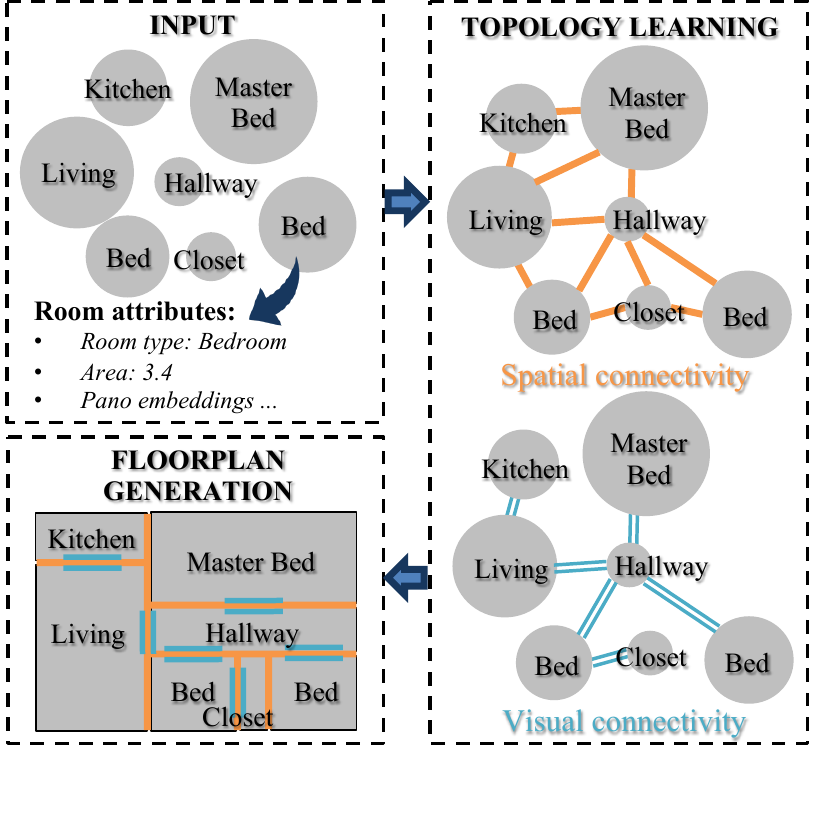}
    \caption{\textbf{Spatial versus visual edges}. Spatial edges are defined by room-room adjacency, while visual connections represent a sub-set of the spatial edges when two rooms share a door or an opening (\eg kitchen is spatially adjacent, but not visually connected to the master bedroom). 
    From room attributes, the proposed method automatically learns both types of topology information (\ie spatial and visual), which can be used in applications like floorplan generation.
    }
    \label{fig:spaVSvis_teaser}
\end{figure}

Motivated by the lack of large-scale floorplan topology information in real estate listing websites (\eg Airbnb, Zillow, Apartments), we propose to extract it using available, or derivable, room-level signals encoded into a graph convolutional network (GCN). Topological information of a floorplan can be easily embedded in a graph, where nodes represent rooms and edges encode relations between them. 
Room adjacency (Fig.~\ref{fig:spaVSvis_teaser}) is characterized as having a {\em spatial} connection; the spatial connection can be either {\em visual} (if there is a door or opening between rooms) or {\em non-visual} (if there is no door or opening). 
In other words, spatial connectivity provides the information about "which room is adjacent to which room", while visual connectivity provides information about "which room connects which room".
To handle the multiple relationships and predict both connection types, we propose the multi-relational GAT as an extension to GAT~\cite{velivckovic2017graph}. Here, we treat each relation as a sub-graph and then fuse the information with a weighted summation of both.

%% Deep Iterative and Adaptive Learning
%To this aim, we propose an iterative and adaptive learning mechanism to progressively predict multiple relationships between rooms with \emph{given attributes} in an indoor environment. During each iteration, we update and generate better node embeddings, which allows for better topological structure learning and vise versa, progressively. We update node embeddings with a weighted message passing mechanism, while using a relational decoder for updating topological structure. It's worth mentioning that conventional similarity metric learning approaches (\eg \cite{chen2020iterative}) cannot simply be applied to our application due to high levels of similarity between our nodes (e.g. two similar non-adjacent bedrooms). Also, current similarity metric learning approaches cannot be extended to multi-relational graphs due to assigning a single score to each pair of nodes. Thus, we learn the graph metric using a relational decoder to make better use of the containment relationship between the spatial and visual edges (i.e. visual edges should be a subset of spatial edges).
We propose an iterative and adaptive learning mechanism to progressively predict multiple relationships between rooms with \emph{given or precomputed attributes} in an indoor environment. Note that conventional similarity metric learning approaches (\eg \cite{chen2020iterative}) cannot simply be applied to our application due to high levels of similarity between our nodes (e.g., two similar non-adjacent bedrooms). Also, current similarity metric learning approaches can be extended to multi-relational graphs only when the graph-structure is available.
Thus, we progressively learn the graph metric using a relational decoder to make better use of the containment relationship between the spatial and visual edges (i.e., visual edges should be a subset of spatial edges).

%Furthermore, we propose an iterative and adaptive graph learning framework to progressively predict multiple relationships of rooms in indoor environments. During iteration, the framework generates better node embedding, which allows to learn a better topological structure of the graph. Then, the enhanced topological structure will further help to generate more informative node embedding. 
%Specifically, node embeddings are updated using a weighted message passing mechanism, while topological structure is updated with a relational decoder. Previous methods (\eg \cite{chen2020iterative}) that model the graph learning problem as similarity metric learning have shown promising performance on knowledge graphs; however, similarity metric learning suitable for knowledge graphs cannot simply be applied to the graph of an indoor space due to their contextual differences. For example, bedrooms in a house usually have similar embeddings; however, they are rarely visually connected, and their spatial connectivity is not guaranteed. Moreover, similarity metric learning can not be applied to multi-relational graphs, since it cannot generate different similarity scores for the same pair of nodes.
%Different from the work in \cite{chen2020iterative} that compute the adjacency matrix based on the similarity of nodes, we learn the graph metric using a relational decoder to make better use of the containment relationship between spatial and visual edges (i.e. visual edges should be a subset of spatial edges).

Our contributions are: 
\begin{itemize}%[leftmargin=*]
    \setlength{\itemsep}{0pt}
    \setlength{\parskip}{0pt}
    \setlength{\parsep}{0pt}
    \item First data-driven method (to our knowledge) to automatically generate floorplan topology using room attributes. Multi-relational (MR) GAT and MR decoder are proposed to model the containment relationship between edges.
    %, which will enable and benefit many applications.
    \item We propose a new iterative learning mechanism to predict multiple relations between graph nodes from scratch, in the absence of any initial topology structure.
    %\item We demonstrate superiority of our proposed iterative topology learning method through extensive experiments. %\sbk{This is a given for all papers.}
    % Furthermore, we demonstrate qualitative and quantitative results on the challenging problem of floorplan generation.
    % Furthermore, qualitative and quantitative results on floorplan generation demonstrates the effectiveness of the proposed method for indoor applications.
    \item We demonstrate the effectiveness of the proposed method for indoor applications (\ie realistic floorplan generation) by employing HouseGAN~\cite{nauata2020house} on the outputs of our method.
    % \item Demonstration of floorplan generation using HouseGAN~\cite{nauata2020house} on the outputs of our method.
\end{itemize}

%\begin{enumerate}
%    \item The first work to automatically generate multiple relations between rooms from room attributes using a data-driven method, which will enable and benefit many applications.
%    \item An iterative graph learning method is proposed to predict multi-relations of rooms (\ie spatial and visual connectivity) from scratch, when the initial graph structure is unavailable.
%    \item Through extensive experiments we demonstrate the superiority of the proposed iterative topology learning method. Furthermore, qualitative and quantitative results on floorplan generation demonstrates the effectiveness and practicality of the proposed method for indoor applications.
%\end{enumerate}

%\sbk{Assumptions?}

%% file: sections/relatedworks.tex
\section{Related Work}

\subsection{Topological relationships of 2D/3D floorplans}
% \noindent\textbf{Topological relationships of 2D/3D floorplans.}
The topological relationships inherent to 2D floorplans have been widely studied in different fields from different perspectives~\cite{goodman2019machine}.
To assist in the task of navigation, Portugal and Rocha~\cite{portugal2012extracting} derived a topological graph-based representation from a grid map of floorplan.
Lin et al.~\cite{lin2017hybrid} merged a topology graph and a grid model to boost the efficiency of spatial analysis and to provide positional information.
Nauata et al.~\cite{nauata2020house} showed that the spatial connectivity of room boosts the performance of floorplan generation task.
In~\cite{hu2020graph2plan}, room connectivity provided by users are used as constraints to generate floorplans. More recently, \cite{armeni20193d} proposed a layered-graph representation, \emph{3D Scene Graph}, to facilitate the holistic understanding of indoor spaces. 

% The topological information used in existing approaches is mainly extracted from grid maps or provided by users.

A popular approach to extract topological relationships is detecting high-level information from grid maps~\cite{thrun1996integrating,joo2010generating,portugal2012extracting,yang2015generation,sithole2018indoor}. The main problem is that grid maps are often unavailable or difficult to obtain in some applications. We argue that it is too restrictive to ask the user or designer to provide such detailed information including both room attributes and connections. This huge drawback can be addressed by extracting topological relationships from room attributes, which is largely unexplored in the literature.
By contrast, we propose to leverage the underlying relational information between rooms to learn topological information from room attributes.
The learned topological information can then facilitate indoor understanding (\eg LayoutGMN~\cite{patil2021layoutgmn}) and more home-related applications such as floorplan generation (\eg HouseGAN~\cite{nauata2020house}, HouseGAN++~\cite{nauata2021house}).

%For example, in floorplan generation tasks, both room attributes and room connections are required by most of the methods~\cite{nauata2020house,hu2020graph2plan}. We argue that it is too restrictive to ask the user or designer to provide such detailed information including both room attributes and connections. Also, room connections cannot be obtained from grid maps, as it is the output of the model. 
%Compared to using grid maps, extracting topological relationships from room attributes is in high demand, but largely unexplored. 
% The missing part is automatically generating room connections from room attributes using data-driven methods.

% \begin{itemize} 
%     \item \href{https://www.mdpi.com/2220-9964/9/5/330}{A Review of Techniques for 3D Reconstruction of Indoor Environments} (section 4.3). A lot of methods about topological modeling (visual and spatial connectivity) are introduced here.
%     \item \href{https://onlinelibrary.wiley.com/doi/abs/10.1111/cgf.14021}{State-of-the-art in Automatic 3D Reconstruction of Structured Indoor Environments} (section 8.3 Room graph computation). It includes some recently published paper about indoor environment modeling using room connectivity.
% \end{itemize}

\subsection{Graph neural networks}
% \noindent\textbf{Graph neural networks.}
GNNs model graph dependencies via message passing between nodes, and have been thoroughly studied~\cite{skarding2020foundations,wu2020comprehensive,zhou2018graph}. 
Recently, GNNs have gained increasing attention in various domains, including knowledge graphs~\cite{schlichtkrull2018modeling,shang2019end,zhu2020collective}, recommendation systems~\cite{ying2018graph,zhang2020gcn}, and other computer vision tasks~\cite{xu2017scene,xue2020learning}.
% (\eg scene representation~\cite{xu2017scene,yang2018graph}, visual relocalization~\cite{xue2020learning}, and fashion compatibility prediction~\cite{cucurull2019context}).
Multi-relational (MR) GNNs  are more intuitive for indoor related tasks, as floorplans typically encode multi-relational connectivity between rooms. To handle multi-relational graphs, R-GCN~\cite{schlichtkrull2018modeling} extends GCN with relation-specific linear transformation.
W-GCN~\cite{shang2019end} utilizes relation-specific learnable weights during information aggregation.
COMP-GCN~\cite{vashishth2019composition} gains extra information by jointly embedding both nodes and relations in a relational graph.

The aforementioned methods are mainly designed for knowledge graph with clean graph structure. However, learning floorplan topology is fundamentally different from these tasks as the graph structure is completely unknown or severely missing. Most of the aforementioned works fail to pass messages along the model when structural information is missing.
To directly adopt these MR-GNNs on floorplan topology learning task, we first need to add initial structures (\eg fully connected graphs, kNN graphs), which will introduce noise to the graph and lead to inaccurate predictions.
Rather than using regular MR-GNNs directly, we extend GAT to multi-relational (MR) GAT for message passing and adopt a MR-decoder to utilize the containment relationship among edge types and learn more precise room connections.

% However, learning floorplan topology is fundamentally different from these tasks as the graph structure is completely unknown or severely missing. Adopting regular GNNs directly by adding naive initial structure (\eg fully connected graphs, kNN graphs) will introduce noise to the graph and lead to inaccurate predictions.

% In addition, floorplans typically encode multi-relational connectivity between rooms. Although methods (\eg~\cite{yu2020generalized,vashishth2019composition,shang2019end,schlichtkrull2018modeling}) have been proposed to deal with multi-relational data, most of them fail to pass messages along the model when structural information is missing.
% Moreover, naive structure initialization methods cannot provide different values for different connection types. 

\subsection{Iterative graph learning}
% \noindent\textbf{Iterative graph learning.}
% In practice, the graph structure is always noisy and incomplete, or may not be available at all.
Most GNN methods ~\cite{velivckovic2017graph,chami2019hyperbolic,kipf2016semi} are only applicable when the graph structure is available and clean. 
To address this limitation, researchers have explored methods~\cite{xue2020learning,franceschi2019learning,grover2019graphite} to automatically generate and iteratively update the graph with (previously) unknown structure.
More recently, \cite{chen2020iterative} cast the graph learning problem as similarity metric learning and proposed to iteratively learn the graph structure and embeddings. Though promising performance was shown on knowledge graphs, it is less effective on floorplan graphs, where connections are not determined only by similarity of two nodes. For example, bedrooms in a house usually have similar embeddings; however, they are rarely visually connected, and their spatial connectivity is not guaranteed.
% Moreover, the current similarity metric learning can not be used for multi-relational graphs, since it cannot generate different values from the same pair of nodes.

%% file: sections/problem.tex
%\section{House topology learning problem}
\section{Problem Definition}

In this paper, our goal is to predict a set of topological relations between the nodes of a graph representing an indoor environment. More specifically, we start with a set of \emph{given} node attributes (room attributes in our case) and the goal is to predict spatial and/or visual connectivity between the given nodes. Our final learned graph embeddings can be used to infer topological structure of indoor environments from scratch, i.e. no links provided, or to complete the topology of partial layouts. 
In this section, we first introduce the dataset we are using, then go over three problem definition including graph structure generation, graph structure completion, and the application of floorplan generation.
% In this section, we first introduce the dataset we are using, then go over the graph construction and finally present potential applications around layout topology completion and floor-plan generation.
%We aim to generate topological relationships (\ie spatial and visual connectivity) of an indoor environment from its entities (\ie room attributes). In this section, we explain the floorplan dataset, graph construction, and possible applications of learned house topology.

%% figure for floorplan sample
\begin{figure}[t]
    \centering
    \includegraphics[trim=0in 0in 0in 0in,clip,width=1.0\linewidth]{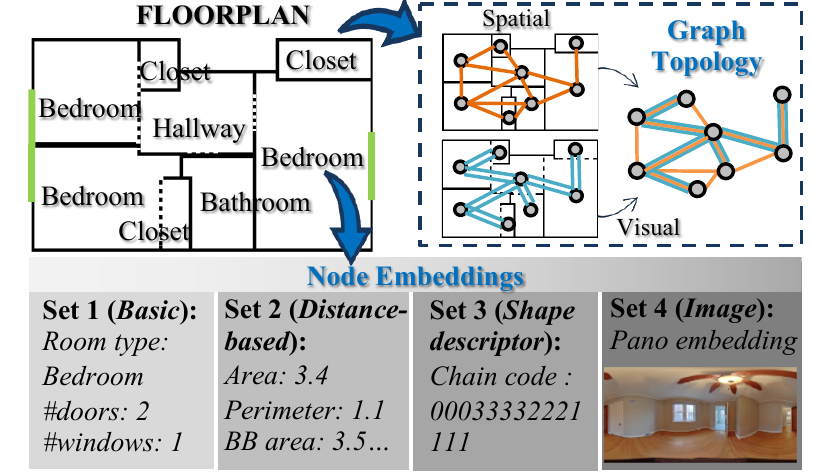}
    \caption{\textbf{Example of floor plan and constructed graph.} Green and dash lines in the floor plan refer to windows and doors, respectively. 
    % The attribute sets are arranged in ascending order of difficulty. 
    Higher-order sets are more specific and harder to get. 
    %Note that good performance can be achieved with the \textit{Basic} attribute set, which improves with more attribute sets (Table~\ref{tab:comp_feat_set}).
    }
    \label{fig:floorplanSample}
\end{figure}

\subsection{Floorplan dataset}
% \noindent\textbf{Floorplan dataset.}
The Zillow Indoor Dataset (ZInD)\footnote{https://github.com/zillow/zind}~\cite{cruz2021zillow} provides 360$^\circ$ panoramas, room layouts, and floorplans collected from 1,575 real unfurnished homes. In total, it contains 2737 floorplans, from which we retrieved 1,217 and constructed graphs containing nodes and multi-relational edges.
The data statistics are summarized in Table \ref{tab:dataStat}. More details (\eg node and edge distributions) of the dataset are provided in the supplementary material.

% \subsection{Graph construction}
% \noindent\textbf{Graph construction.}
We then represent each floorplan as a graph.
% Floorplans can naturally be represented as a graph. 
Each node in a floorplan represents a room, and edges represent the connectivity between two rooms. 
We define two types of edges, spatial and visual (Fig. \ref{fig:spaVSvis}). Spatial edges are defined by room-room adjacency in a floorplan, while visual connections represent a sub-set of the spatial edges when two rooms share a door or an opening.
% ~Fig. \ref{fig:floorplanSample}.
%Four types of features are extracted for each room based on normalized floorplans and panoramas: (1) basic room information, (2) distance-based features, (3) shape descriptors, (4) image features. The contribution and effect of different types of features are tested in the experiment section.
We represent nodes using four types of room features:
\begin{itemize}%[leftmargin=*]
    \setlength{\itemsep}{0pt}
    \setlength{\parskip}{0pt}
    \setlength{\parsep}{0pt}
    \item \textbf{Basic room information} (Set~1): room type, number of doors, windows, and openings of the room. 
    \item \textbf{Distance-based features} (Set~2): perimeter and area of the room, maximum length and width of walls, and ratio between room area and its bounding box area.
    \item \textbf{Shape descriptors} (Set~3): we use the chain code~\cite{freeman1961encoding} to represent the shape of a room. The chain code is extracted by traversing the boundary of the object and recording the direction of each step.
    \item \textbf{Image features} (Set~4): extracted from panoramas using embeddings from a scene classification model~\cite{cruz2021zillow}.
\end{itemize}

% \textbf{Basic room information} (Set~1): room type, number of doors, windows, and openings of the room. 
% \textbf{Distance-based features} (Set~2): perimeter and area of the room, maximum length and width of walls, and ratio between room area and its bounding box area. 
% \textbf{Shape descriptors} (Set~3): we use the chain code~\cite{freeman1961encoding} to represent the shape of a room. The chain code is extracted by traversing the boundary of the object and recording the direction of each step.
% %We use the chain code~\cite{freeman1961encoding} as shape descriptors to represent the shape of rooms. The idea of chain code is to traverse the boundary of the object, and record the direction of each step. 
% \textbf{Image features} (Set~4): extracted from panoramas using embeddings from a scene classification model~\cite{cruz2021zillow}.

Note that all room attributes are invariant to translation and rotation. Features in Sets 1 and 4 are also invariant to scale, while the other two are not.
Fig.~\ref{fig:floorplanSample} shows an example of a floorplan, input attributes and constructed graphs from the dataset. Detailed description of topology and attribute extraction are in the supplementary material. 
In our work, we assume all these room-level attributes are given. This is the input to our system, and our goal is to predict the layout. In practice, these room attributes can be derived from the panorama images, either manually (\eg~\cite{cruz2021zillow, armeni2017joint}) or automatically from SOTA algorithms for scene classification (\eg~\cite{liu2019indoor, ismail2018understanding}) and room layout estimation (\eg~\cite{cabral2014piecewise,pintore2019automatic,yang2019dula}). In Table~\ref{tab:comp_feat_set} we study the effect of the different attributes.

%% table for data statistics
\begin{table}[t]
\caption{\textbf{Data statistics.}}
\vspace{-2mm}
\begin{center}
\begin{tabular}{c | c c c}
\toprule
  & Total & Avg & (Min, Max)\\
\hline
\# of graphs & 1217 & --  & -- \\
\# of nodes & 12253 & 10.07  & (2, 30) \\
\# of spatial edges & 55242 & 45.39  & (8, 144) \\
\# of visual edges & 12138 & 9.97  & (2, 32) \\
\bottomrule
\end{tabular}
\end{center}
\label{tab:dataStat}
\end{table}

\subsection{Graph structure generation}
The goal of graph structure generation is to predict different types of connections (\ie visual, wall, and spatial connections) given only room attributes and without any structure information. The adjacency matrix of the input graph is an identity matrix, meaning that each room is connected to itself and not to any other rooms in the floorplan. The dataset is split across graphs. We use 700 floorplan graphs for training, 200 floorplans for validation, and the rest (\ie, 317 floorplans) for testing. All edges in training graphs are in the training set, and the model is asked to predict unseen links (\ie all edges) for test graphs.

\subsection{Graph structure completion}
Given room attributes and partial spatial structure information, the goal is to complete the spatial connection (\ie the rooms are connected or not) and predict the specific type of the connection (\ie the rooms are connected by door (visual) or wall). 
Let $A$ be the adjacency matrix of the graph generated from original spatial edges of a floor plan; we randomly remove a subset of spatial edges (\eg 20\% - 80\% of edges) and use the remaining observations to generate an incomplete adjacency matrix $\bar{A}$. We use the same data split as graph structure generation task (\ie 700 graphs for training, 200 for validation and 316 for testing). All edges in training graphs are used for training. We then evaluate on a fixed set of 20\% of connections that was not observed.

\begin{figure*}[t]
    \centering
    \includegraphics[trim=0in 0in 0in 0in,clip,width=1\linewidth]{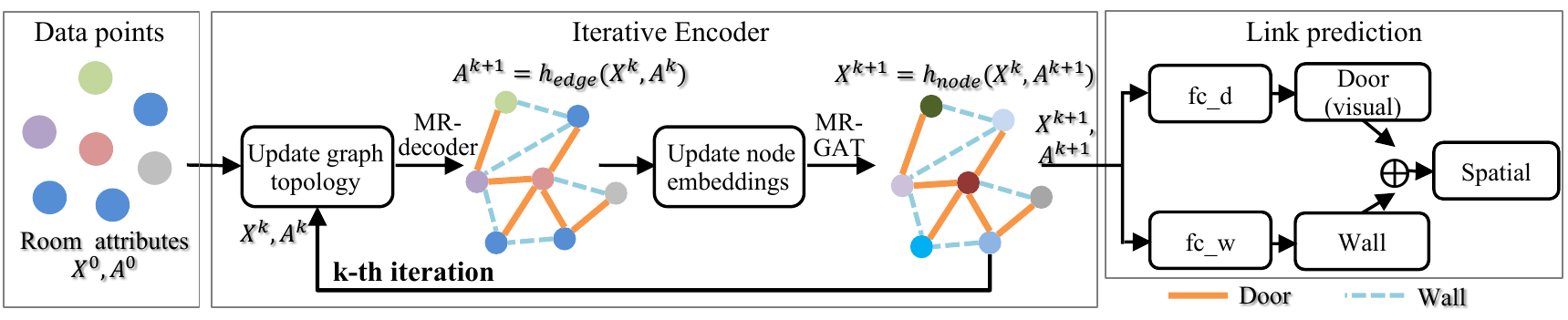}
    \caption{The architecture of the proposed model. The model is optimized with binary cross entropy loss on positive and negative edges of all visual, wall, and spatial connectivity.}
    \label{fig:architecture}
\end{figure*}

% \subsection{Floorplan generation}
\subsection{Indoor application using generated topology}
% \subsection{Applications}
% \noindent\textbf{Applications.}
The proposed learned representation includes rich information about an indoor environment. We show how this topological information can benefit the challenging task of home layout generation, from room-level attributes.
As demonstration, we use HouseGAN~\cite{nauata2020house}, which takes in room attributes and room spatial information.
We replace the ground truth (GT) room connections with our estimated graph topology, and show that it can generate compatible results with using the GT. 
Following~\cite{nauata2020house}, results are evaluated in terms of realism, diversity, and compatibility. Qualitative result shows the realism of generate floorplan. Diversity and compatibility are measured by the Frechet Inception Distance (FID) scores and graph edit distance~\cite{sanfeliu1983distance} (GED), respectively.
We show a significant performance boost on HouseGAN~\cite{nauata2020house} by adopting the topological information generated from the proposed method.
Note that we do not train end-to-end, as HouseGAN only includes a single edge type (\ie spatial edges), whereas our model predicts both spatial and visual connectivity. 
For floorplan generation task, we fine-tune the pre-trained HouseGAN model with our dataset and follow the same setting as in HouseGAN. Since the room types in HouseGAN are different from ours, we map our room types to match theirs.

% %% figure for x-alignment
% \begin{figure}[t]
%     \centering
%     \includegraphics[trim=0in 0in 0in 0in,clip,width=0.8\linewidth]{figures/node_dist.png}
%     \caption{Node distribution}
%     \label{fig:node_dist}
% \end{figure}

% %% figure for x-alignment
% \begin{figure}[t]
%     \centering
%     \includegraphics[trim=0in 0in 0in 0in,clip,width=0.8\linewidth]{figures/edge_dist.png}
%     \caption{Edge distribution}
%     \label{fig:edge_dist}
% \end{figure}

%% file: sections/method.tex
\section{Methodology}
The goal of this paper is to use relational information between graph nodes to predict its topology, when very little to no topological information is available. By comparison, traditional approaches assume a percentage of edges are available and aim to predict the rest.
% \Naji{\textbf{Yu: can you please verify my understanding of literature in this sentence is accurate and correct?}}\Yu{Naji: Yes, that's correct. We also do experiments when a percentage of edges is given.}
%With this paper we address the challenging scenario where a graph topology is either incomplete or completely missing. We proposed to leverage the underlying relational information between rooms in an indoor space to extract topological information from room attributes. 
The architecture of our proposed model is shown in Fig.~\ref{fig:architecture}. Our architecture allows us to jointly predict both visual and spatial edges.
We first generate an incomplete graph with an identity adjacency matrix from a floor plan.
%Firstly, from a floor plan we generate an incomplete graph with identity adjacent matrix. 
The incomplete graph is then fed into an iterative encoder, which recursively updates graph topology and node embeddings. 
With node embeddings as the input, the multi-relational (MR) decoder then learns more precise room connections by leveraging the containment relationship between spatial and visual edges (\ie visual edges should be a subset of spatial edges).

Let $G = (V, E)$ be an undirected graph, where $V$ is a set of $N$ nodes with node attributes $X = \{x_1, x_2, ..., x_N\}, x_i \in \mathbb{R}^{1\times F}$. $F$ is the feature dimension of each node.
The true adjacency matrix $A \in \mathbb{R}^{N \times N \times R}$ is formulated from edges $E$, where $R$ is the number of relation types in the graph. 
Suppose the initial adjacency matrix $A^0$ has missing or incomplete information. 
The objective of the model is to complete the graph topology by exploiting potential relationships based on room attributes.
Specifically, an iterative encoding function is applied to transform the initial features $X^0$ to a new feature representation.
The multi-relational decoding function, which takes the new feature representations as input, is then used to predict the probability of whether or not an edge exists between pairs of nodes $(v_i, v_j)$, as well as the edge type.

We predict multiple types of edges (\ie spatial and visual edges). Since visual edges are a subset of spatial edges, we split spatial edges into two exclusive sets: door (\ie visual) connectivity and wall connectivity. We then jointly predict all three types of edges to fully explore and utilize the containment relationship among them.

\subsection{Iterative encoder}
The iterative encoder recursively augments the initial graph topology and node embeddings to improve the performance of downstream prediction tasks.
The initial node embeddings $X^{0}$ and adjacency matrix $A^{0}$ are the input of our iterative encoder. $X^{0}$ is the original node (room) attributes, and $A^{0}$ is set to either an identity matrix or an incomplete adjacency matrix.
In each iteration $k$, we first update the graph topology from $A^{k}\in\mathbb{R}^{N\times N}$ to $A^{k+1}\in\mathbb{R}^{N\times N}$ using a multi-relational scoring function, then compute new node embeddings $X^{k+1}\in\mathbb{R}^{N\times F^\prime}$ based on $X^{k}\in\mathbb{R}^{N\times F}$ and $A^{k+1}$. Inspired by~\cite{shang2019end}, we extend GAT~\cite{velivckovic2017graph} to multi-relation GAT (MR-GAT) for message passing, where we treat each relation as a subgraph and then conduct weighted summation on all the subgraphs. The process of each iteration block can be expressed as:
\begin{equation}
\begin{aligned}
A^{k+1} =& ~ h_{edge} (X^{k}, A^{k}), \\
X^{k+1} =& ~ h_{node} (X^{k}, A^{k+1}),
\end{aligned}
\end{equation}
where $h_{edge}$ and $h_{node}$ denote the function to update graph topology and node embeddings, respectively.

% ($X^{k}\in\mathbb{R}^F$ and $ X^{k+1}\in\mathbb{R}^{F^\prime}$)
%  In this work, the encoder is base Graph attention network (GAT), and the decoders are fully connected layers to predict the connectivity score between pairs of nodes $(i, j)$. 

\subsubsection{Updating graph topology}
% \noindent\textbf{Updating graph topology.}
Previous approaches (\eg~\cite{chen2020iterative, kipf2016variational}) that model the graph structure learning problem as similarity metric learning have shown promising performance on knowledge graphs. However, these methods are unable to cope with the floorplan graphs, in which connections are not determined only by the similarity of two nodes. 
Moreover, similarity metric learning can only generate a single matrix to represent one type of room relationship given the node embeddings, and hence cannot be applied to multi-relational graphs.
Instead, we use a multi-relational scoring function to make better use of the containment relationship between spatial and visual edges. Specifically, we decompose spatial connectivity into door (visual) and wall connectivity and predict them separately. 
Then spatial edges can be calculated by summing up door and wall edges. The prediction of door and wall edges can further be restricted by the binary cross-entropy loss applied on spatial edges. 
Therefore, the probability $a^{k}_{i,j,r}$ that node $i$ and $j$ are connected by relation $r$ (\ie door, wall, or spatial) at the $k$-{th} iteration is updated by:
\begin{equation}
a^{k+1}_{i,j,r}= 
\begin{cases}
    f \left( [\vec{x}_i^{k}, a_{i,j,r}^{k}, \vec{x}_j^{k}] \right), & r\in \{door, wall\}\\
    a^{k+1}_{i,j,door} + a^{k+1}_{i,j,wall}, & r\in \{spatial\}
    \end{cases}  
\end{equation}
where $x_i$ and $x_j$ are the embeddings of node $i$ and $j$. The scoring function $f(\cdot)$ is implemented with three fully-connected layers.

% \begin{equation}
% A^{k+1}= \left[A_{door}^{k+1};~
% A_{wall}^{k+1};~
% A_{door}^{k+1}+A_{wall}^{k+1}\right]
% % \begin{bmatrix}
% % A_{door}^{k+1}\\
% % A_{wall}^{k+1}\\
% % A_{door}^{k+1}+A_{wall}^{k+1}
% % \end{bmatrix}
% \end{equation}

\begin{figure}[t]
    \centering
    \includegraphics[trim=0in 0in 0in 0in,clip,width=1\linewidth]{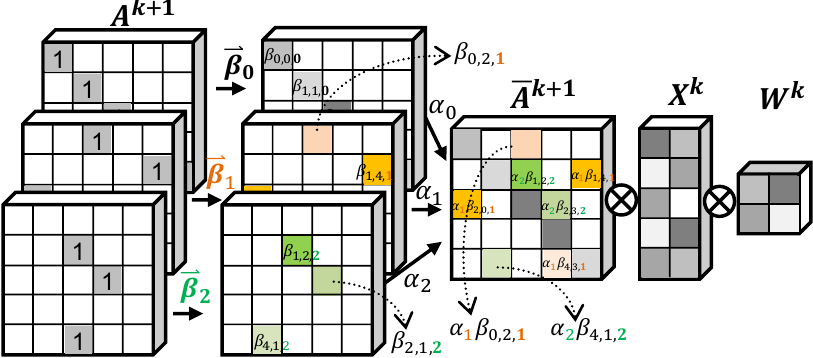}
    \caption{\textbf{Multi-relational GAT.} Relation is indicated by color. We apply an attention mechanism on each subgraph of different relations, then perform weight summation on the output of subgraphs. MR-GAT determines the weights of each subgraph when fusing the GAT embeddings for a node. 
    }
    \label{fig:mrgat}
\end{figure}

\subsubsection{Updating node embeddings}
% \noindent\textbf{Updating node embeddings.}
MR-GAT is employed to handle multi-relation message passing, which measures the various types of relationships differently during aggregation and learns the weights adaptively during network training. 
Specifically, the encoder treats a multi-relation floorplan graph as multiple single-relation subgraphs, where each subgraph represents a specific type of connection (\ie door or wall connection). For each subgraph with relation $r$, we then conduct self-attention on the nodes to indicate the importance of node $j$’s features to node $i$. We first apply an attention mechanism $g$, which is the same as~\cite{velivckovic2017graph}, on the concatenated features of node $i$ and $j$ to compute the attention coefficient, then use the softmax function to normalize it across all choices of $j$ in each relation $r$. The attention coefficients can be expressed as:
\begin{equation}
\begin{aligned}
\beta_{i,j,r} =&\text{softmax}_{j,r} \left( g([\vec{x}_i, \vec{x}_j])  \right)\\
=&  \frac{ \text{exp}\left( g([\vec{x}_i, \vec{x}_j])\right)}{\sum_{j\in \textbf{N}(i,r)}  \text{exp}\left( g([\vec{x}_i, \vec{x}_j])  \right)},
\end{aligned}
\end{equation}
where [$\cdot$, $\cdot$] represents concatenation operation, and $\textbf{N}(i,r)$ denotes sets of rooms that are connected with node $i$ by relation $r$ (\ie wall or door connection). Hence the output of each subgraph can be expressed as:
\begin{equation}
\vec{x}_{i,r}^\prime =  \sum_{j\in \textbf{N}(i,r)} \beta_{i,j,r} \vec{x}_j W_r,
\end{equation}
where $W_r$ is the weight matrix of linear transformation for relation $r\in \{wall, door\}$.

Then we extend GAT to MR-GAT by performing weight summation on subgraphs of different relations $r$. The MR-GAT determines how much weights to give to each subgraph when fusing the GAT embeddings for a node. Hence, the embeddings of node $i$ can be updated by:
\begin{equation}
{\vec{x}_i}^\prime= \sigma \left(  \sum_{r} \alpha_r  \sum_{j\in \textbf{N}(i,r)} \beta_{i,j,r} \vec{x}_j W_r  \right),
\end{equation}
where $\alpha_r$ denotes the attention coefficient for each subgraph $r$, and is learnt during training. 
As shown in Fig.~\ref{fig:mrgat}, this process is essentially matrix multiplication.
For each subgraph ${A}^{k+1}$, we first conduct self-attention on the nodes. 
Then the attention-weighted adjacency matrix $\bar{A}^{k+1}$ will be the weighted sum of the output subgraphs, and can be written as:
\begin{equation}
\bar{A}^{k+1}=  \sum_{r} \alpha_r \vec{\beta_{r}} A^{k+1}.
\end{equation}
In matrix form, the node embedding update can be expressed as:
\begin{equation}
X^{k+1}= \sigma \left( \bar{A}^{k+1} X^{k} W^k \right).
\end{equation}

\subsection{Link prediction decoder}
Similarly, we take advantage of the containment relationship between spatial and visual edges and apply the same MR decoder for the link prediction task by decomposing spatial connectivity into door and wall connectivity, and then predicting them separately.
After the iterative encoder, the model splits into two decoder branches, which predict door and wall edges separately. The two decoders have the same structure, each containing three fully-connected layers. Then, the door and wall predictions are fused to obtain the probability of whether two nodes are connected (\ie spatial edges).
Furthermore, we found that the initial features $X^0$, which contain node-wise room properties (\eg room shape), help to improve the prediction performance. Hence, we add a long skip connection by concatenating the initial features $X^0$ and the output of the encoder $X^{k+1}$. Hence, the probability that two nodes are connected by relation $r$, $(r\in \{spatial, door, wall\})$, can be computed by

$${p_r} = {mr\_dec}([X^{k+1},X^{0}], A^{k+1}), $$
    	
The whole process of our method is described in Alg.~\ref{alg:pyramid_fusion}.

% Then the model splits into two branches. The first branch will predict door connections, which is also visual connections. The second branch will predict wall connections. Then the door and wall predictions be fused by a fully connected layer to predict spatial edges. 

% Furthermore, we find that the initial feature $X$ itself contains some rooms properties (\eg room shape), which also helps to improve the link prediction performance. Hence, we add a long skip connection by concatenating the initial features $X$ and the output of graph attention layers $\hat{X}$. The final output of encoder would be $H = [X, g(A, X)]$.

% The decoders output the probability that two nodes are connected by an edge. The proposed model contains three decoders for predicting visual, wall, and spatial connectivities, respectively. The two decoders $d_1$ and $d_2$ for visual and wall edges prediction have the same structure. Each of them contains two fully connected layers. Hence, the predicted outputs for visual and wall edges are $P_1 = d_1(H)$, and $P_2 = d_2(H)$, respectively.
% Then the visual and wall edge predictions are fused by a fully connected layer to predict spatial edges. $P_1$ and $P_2$ are concatenate and feed in to the other decoder $d_3$. The predicted outputs for spatial edges are $P_3 = d_3([P_1, P_2])$.

\begin{algorithm}[t]
\SetKwInput{KwInput}{Input}                % Set the Input
\SetKwInput{KwOutput}{Output}              % set the Output
\DontPrintSemicolon
  \KwInput{Node embeddings ${X^0}$, incomplete or unavailable adjacency matrix ${A^0}$.}
  \KwOutput{Predicting multi-relations of the floorplan graph $p_{spatial},p_{door},p_{wall}$.}
  \BlankLine
  ${A^0}$ = identity matrix $I$ if ${A^0}$ is not available
  
%   \tcc{Until $K$ iterations.}
    \While{$k\leq K$ }{
        \tcc{Updating graph topology}
        \eIf{$r\in\{door, wall\}$}
        {$A_r^{k+1}=f(X^{k},{A}^{k})$}
        {$A_r^{k+1}=A_{door}^{k+1}+A_{wall}^{k+1}$}
        $A^{k+1}=\left[A_{spatial}^{k+1}; A_{door}^{k+1}; A_{wall}^{k+1}\right]$
        
        \tcc{Updating node embeddings}
        $~\bar{A}^{k+1}=  \sum_{r} \alpha_r \vec{\beta_{r}} A^{k+1}$
    	$X^{k+1}=\sigma(\bar{A}^{k+1}{X}^{k}{W}^{k})$
  }
  \tcc{Downstream task (link prediction)}
    $p_{spatial},p_{door},p_{wall} = {mr\_dec}(concat([X^{k+1},X^{0}]),A^{k+1})$
 \caption{Iterative learning of the multi-relational topology of a floor plan graph.}
 \label{alg:pyramid_fusion}
\end{algorithm}

\subsection{Loss function}
We jointly predict all three types of edges by combining the loss function of each task at the end of every iteration during training.
Therefore, the loss function $\mathcal{L}$ of the proposed model can be expressed as:
\begin{align}\label{eq:loss}
\mathcal{L} &=  \sum_{i=1}^{K} (\mathcal{L}_{vis}^{(k)} + \lambda_1 \mathcal{L}_{wall}^{(k)}  + \lambda_2 \mathcal{L}_{spa}^{(k)})
\end{align}
where $\mathcal{L}_{vis}$, $\mathcal{L}_{wall}$, and $\mathcal{L}_{spa}$ denote the binary cross-entropy loss for visual (door), wall, and spatial connectivity, respectively. $\lambda_1$ and $\lambda_2$ are non-negative hyperparameters. We set $\lambda_1=\lambda_2=1$ in our experiments. Sensitivity analysis for these two hyper-parameter is shown in the Supplementary material.
% \Ivo{Yu: we need to provide the values and mention that sensitivity analysis on those will be in the supplementary.}

%% file: sections/experiments.tex
\section{Experiments}
To evaluate our framework, we compare it with a variety of baselines on graph structure generation and completion. We further report results on floorplan generation as one of the applications of the proposed method. 
Ablation studies are conducted to show the contribution of each attribute set, as well as each component of our method and its limitations.
% We also perform ablation studies to show the contribution of each attribute set, as well as each component of our framework and its limitations.
% \Will{This is great. The tense is typically present though, right? e.g. "we comprehensively compare" and "we further report"}
%We comprehensively evaluate our method by comparing it with a variety of networks in terms of graph structure generation and completion. Furthermore, we demonstrate the effectiveness and practicality of our model on indoor applications (\ie floorplan generation).
%Finally, we conduct an ablation study as a deep-dive revealing the effect of various feature sets, the contributions of the components introduced in this work, as well as the limitations of the work.
Implementation details are described in the supplementary material.

\subsection{Experimental settings}
%\subsubsection{Evaluation protocol}
% In this section, we describe the experimental setting as well as evaluation protocols.
%We now review the evaluation protocol of various tasks.

\noindent\textbf{Graph structure generation.}
We predict different types of connections (\ie visual, wall, and spatial connections) given only room attributes and without any structure information. Following previous works~\cite{kipf2016semi,chami2019hyperbolic,chen2020iterative}, we evaluate link prediction by measuring area under the receiver operator characteristic (ROC) curve (AUC) and average precision (AP) on the test set.

%% table: graph generation from scratch
\begin{table}[t!]
    \caption{\textbf{Generating topology from scratch.} We report results with ROC AUC and AP.}
    {\begin{center}
    \begin{adjustbox}{max width=1\linewidth}
        \begin{tabular}{c ccc c ccc}
            \toprule
            & \multicolumn{3}{c}{AUC} && \multicolumn{3}{c}{AP}  \tabularnewline 
            \cline{2-4} \cline{6-8} 
            & Spatial & Wall & Door && Spatial & Wall & Door \tabularnewline
            \toprule
            GCN$_{\text{kNN}}$~\cite{kipf2016semi} & 73.10 & 54.72 & 82.23 && 73.80 & 54.10 & 80.08 \tabularnewline
            GAT$_{\text{kNN}}$~\cite{velivckovic2017graph} & 73.67 & 54.46 & 83.22 && 74.54 & 54.82 & 83.39 \tabularnewline
            HGCN$_{\text{kNN}}$~\cite{chami2019hyperbolic} &73.96 & 54.99 & 86.90 && 76.41 & 55.80 & 86.70 \tabularnewline
            MLP & 73.93 & 56.74 & 83.41 && 74.25 & 55.06 & 81.41 \tabularnewline
            HNN~\cite{ganea2018hyperbolic} & 75.21 & 58.33 & 86.27 && 76.59 & 56.87 & 86.23 \tabularnewline
            IDGL+{sim}~\cite{chen2020iterative} & 72.19 & 56.81 & 83.48 && 73.00 & 56.01 & 81.08 \tabularnewline
            IDGL+{dec}~\cite{chen2020iterative} & 79.23 & 63.63 & 89.42 && 79.48 & 63.89 & 87.13 \tabularnewline
            Ours & \textbf{81.64} & \textbf{65.67} & \textbf{91.73} && \textbf{81.86} & \textbf{65.36} & \textbf{91.31} \tabularnewline
            \bottomrule
        \end{tabular}
    \label{tab:graphGeneration}
    \end{adjustbox}
    \end{center}
    }
\end{table}

% %% figure: graph completion (completion)
% \begin{figure*}[t]
%     \centering
%     \includegraphics[trim=0in 0in 0in 0in,clip,width=0.9\linewidth]{figures/completion.pdf}
%     \caption{\textbf{Topology completion results.} Given different percentage of observations (\ie spatial connections), we report results on a fixed set of 20\% of spatial connections that was not observed. Predictions of the specific connection type (\ie the door and wall connections) are also reported with AUC.}
%     \label{fig:completion}
% \end{figure*}

%% figure: graph completion (completion)
\begin{figure}[t]
    \centering
    \includegraphics[trim=0in 0in 0in 0in,clip,width=1\linewidth]{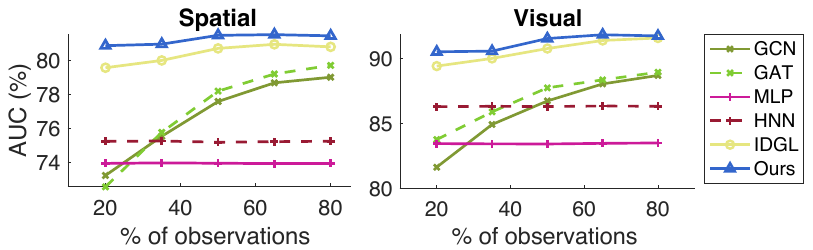}
    \caption{\textbf{Topology completion results.} Given different percentage of observations (\ie spatial edges), we report results on a fixed set of 20\% of spatial edges that was not observed. Predictions of the specific connection type (\ie visual) are also reported.}
    \label{fig:completion}
\end{figure}

\noindent\textbf{Graph structure completion.}
Given room attributes and partial spatial edges (\eg 20\% - 80\% of edges), the goal is to complete the spatial connection (\ie the rooms are connected or not) and predict the specific type of the connection (\ie the rooms are connected by door (visual) or wall). 
We evaluate on a fixed set of 20\% of connections that was not observed. As with the graph structure generation task, results are reported with ROC AUC and AP.

\noindent\textbf{Floorplan generation.}
Floorplan generation is a challenging task that highly depends on topological information. We selected the problem of house layout generation and showed how representations learned using our method can improve the state-of-the-art in this task, both qualitatively and quantitatively.
Specifically, we fine-tune the pre-trained HouseGAN~\cite{nauata2020house} model with our dataset to generate floorplans. Room attributes and room spatial information are the input of HouseGAN. We replace the ground truth (GT) room connections with our estimated graph topology, and show that it can generate compatible results with using the GT. We follow the same setting as in HouseGAN, and report with both qualitative and quantitative results (\ie the Frechet Inception Distance (FID) scores and graph edit distance~\cite{sanfeliu1983distance} (GED)).

%%The proposed method provides the basis (\ie the topological information) for lots of indoor applications (\eg house layout generation).
%%To demonstrate the effectiveness and practicality of our method, we qualitatively and quantitatively compare the floorplan generation performance with and without our estimated topology information.
% As demonstration, we use HouseGAN~\cite{nauata2020house}, which takes in room attributes and room spatial information. We replace the ground truth (GT) room connections with our estimated graph topology, and show that it can generate compatible results with using the GT. Following~\cite{nauata2020house}, results are evaluated in terms of realism, diversity, and compatibility. Qualitative result shows the realism of generate floorplan. Diversity and compatibility are measured by the Frechet Inception Distance (FID) scores and graph edit distance~\cite{sanfeliu1983distance} (GED), respectively.

%% figure: housegan
\begin{figure*}[t]
    \centering
    \includegraphics[trim=0in 0in 0in 0in,clip,width=0.75\linewidth]{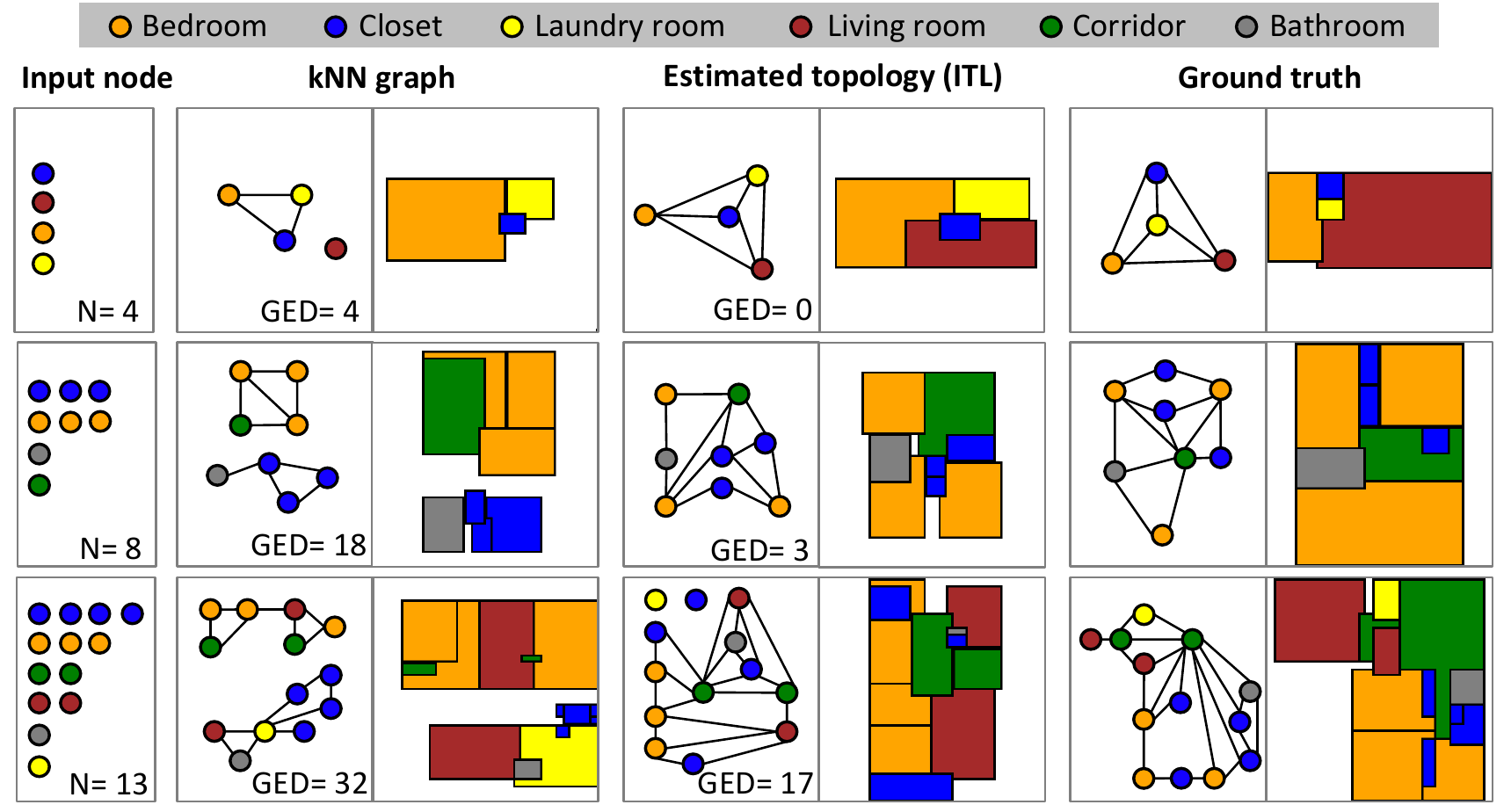}
    \caption{\textbf{Floorplan generation with different input topology}. We compare the output of HouseGAN given different connectivity information (\ie kNN graph and estimated topology using ITL). The graph topology generated by our method is more faithful to GT, and has lower graph edit distance (GED). For complicated cases (\ie node $N>=10$), ITL can still provide reasonable topology to HouseGAN for generating realistic floorplans. More examples are in the supplementary material.
    }
    \label{fig:housegan}
\end{figure*}

\subsection{Graph structure generation}
Table~\ref{tab:graphGeneration} reports the performance of our method in comparison to other baselines on graph structure generation. 
For state-of-the-art GNN models, we consider GCN~\cite{kipf2016semi}, GAT~\cite{velivckovic2017graph}, and Hyperbolic Convolutional Networks (HGCN)~\cite{chami2019hyperbolic}. 
We also consider feature-based approaches: Multilayer Perceptron  (MLP) and its hyperbolic variant (HNN)~\cite{ganea2018hyperbolic}, which do not utilize the graph structure.
Moreover, we compare with IDGL~\cite{chen2020iterative} and its variant, which are iterative graph learning methods. We use the same decoder for down-stream prediction task for all the methods.

\textbf{GCN}, \textbf{GAT}, and \textbf{HGCN} need graph structure information to perform message passing. 
Since the graph topology is inaccessible, we adopt the kNN graph to generate the initial topology and build various GNN$_{\text{kNN}}$ baselines (\ie GCN$_{\text{kNN}}$, GAT${_\text{kNN}}$, HGCN$_{\text{kNN}}$). The kNN graph is computed based on node attributes, and is constructed during preprocessing before applying a GNN model. We can see that these methods do not perform well when the true graph topology is inaccessible and GNNs are not directly applicable (Table~\ref{tab:graphGeneration}).
Note that most of existing multi-relational graph methods (\eg RGCN~\cite{schlichtkrull2018modeling}, WGCN~\cite{shang2019end}, COMP-GCN~\cite{vashishth2019composition}) fail to handle the task here due to missing of initial graph structure for all relations. Like GCN and GAT, initial graph structure is needed for multi-relational graph methods to pass messages along the model. However, kNN can not provide topological information for different edge types, since the same node embeddings are given. Therefore, we do not compare with those methods.

%% table: houseGAN
\begin{table}[t]
\caption{\textbf{Quantitative results}. We evaluate the diversity and compatibility of HouseGAN with different connectivity information.
%(\ie fully connected graph, kNN graph, estimated topology using our method, and the ground truth connections). 
Note: the lower the FID, GED scores, the better.
}\label{tab:housegan}
\begin{center}
\begin{adjustbox}{max width=1\linewidth}
    \begin{tabular}{c | c c}
        \toprule
         & Diversity (FID) & Compatibility (GED)\\
        \hline
        Fully connected graph & 83.79 & 8.64  \\
        kNN graph & 33.56 & 6.53 \\
        Estimated topology & 19.68 & 4.71 \\
        Ground truth & 17.54 & 4.22 \\
        \bottomrule
    \end{tabular}
\end{adjustbox}
\end{center}
\end{table}

\textbf{MLP} and \textbf{HNN} are feature-based methods and do not need neighbor information. As shown in Table~\ref{tab:graphGeneration}, feature-based methods (\ie MLP and HNN) may perform better than GNN models when the true graph topology is unavailable. One possible reason is that the graph structure initialization step used in GNNs is inaccurate and introduces a significant amount of noise. 

\textbf{IDGL} iteratively learns graph structure and graph embedding during training. The first IDGL baseline (\ie IDGL+sim) casts the graph structure learning problem as a similarity metric learning problem. For the baseline of IDGL+dec, we replace similarity metric learning with our proposed topology learning method to update graph structure. Since IDGL cannot handle multi-relation graphs, we only consider spatial edges when updating graph structures. Compare to IDGL baselines, our method consistently achieves better performance on all types of edges.

\subsection{Graph structure completion}
The results on graph structure completion task are shown in Figure~\ref{fig:completion}.
Although feature-based methods (\ie MLP, HNN) perform better than GNN models (\ie GCN, GAT) when the graph structure is either unknown or severely missing, its performance remains almost unchanged even if more neighbor information is given. 
With more observations, GNN models tend to outperform feature-based methods. 
Iterative learning methods (\ie IDGL and ours) achieve better results than GNNs and feature-based methods by a large margin and have stable performance with different proportion of observations.
Note that the IDGL model we compare with is IDGL+dec, which uses the same topology updating method as ours. 
Compared to IDGL, our method consistently achieves much better results by adopting a multi-relational message passing mechanism.

\subsection{Indoor application using generated topology}
Figure~\ref{fig:housegan} shows the qualitative comparison of different graph structure (\ie kNN graph and ITL graph) usage for floorplan generation. We show that ITL can generate effective and practical floorplan topological information with much lower graph edit distances, which leads to more realistic floorplan generation using HouseGAN~\cite{nauata2020house}.
The kNN graph fails to provide compelling topology for HouseGAN, since it predicts connections based on the similarity of room attributes. Rooms generated with kNN graphs tend to be clustered by the room type (\eg bedrooms are connected to bedrooms, closets are connected to closets).

We demonstrate the importance of topological information for house layout generation tasks in Table~\ref{tab:housegan}.
Compared to fully connected graph and kNN graph, we observe significant performance boost on HouseGAN by adopting the topological information estimated from ITL. The floorplans generated with ITL are very close to the ones generated with GT graphs in terms of diversity and compatibility, which also shows ITL can provide reasonable topology for indoor applications, and can well replace GT graphics.

%% tab: ablation
\begin{table}[t]
\caption{\textbf{Ablation study.} We measure the contributions of iterative learning scheme, multi-relation (MR) GAT, and MR prediction.}\label{tab:ablation}
\begin{center}
\begin{adjustbox}{max width=0.8\linewidth}
\begin{tabular}{c | c c c}
\toprule
  & Spatial & Wall & Door\\
\hline
ITL & \textbf{81.64} & \textbf{65.67} & \textbf{91.73} \\
w/o IL & 75.62 & 57.92 & 88.61 \\
w/o MR-GAT & 80.93 & 65.15 & 90.46 \\
w/o MR decoder & 78.43 & 61.38 & 89.43 \\
\bottomrule
\end{tabular}
\end{adjustbox}
\end{center}
\end{table}

%% figure: num. of iteration
\begin{figure}[t]
    \centering
    \includegraphics[trim=0in 0in 0in 0in,clip,width=0.9\linewidth]{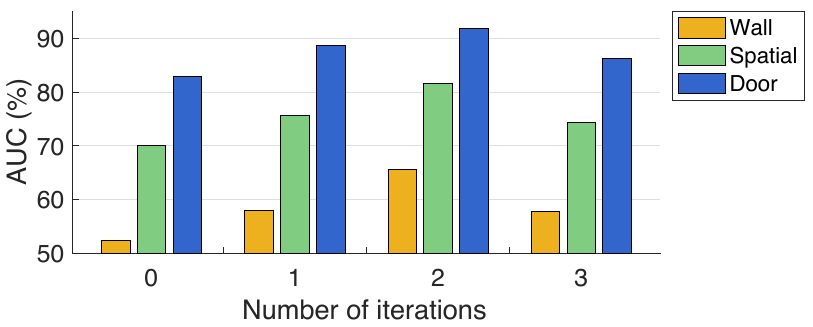}
    \caption{\textbf{Results with different numbers of training iterations.} Note that the model with zero iteration directly feeds the node features to the decoder and predicts links.}
    \label{fig:iter}
\end{figure}

\subsection{Ablation study}
% In this section, we discuss the results of an ablation study, the effect of iteration, and the importance of various room attributes. 
%We will conduct an ablation study to gain insight into the role of the components introduced in the work, model analysis, and the effect of various feature sets.
%We also visualize the graph topology learned by ITL as well as some failure cases, and conduct hyper-parameter analysis in Supplementary material.

\subsubsection{Contribution of each component} 
% \noindent\textbf{Ablation study.} 
%The ablation analysis findings for various modules in our models are shown in Table~\ref{tab:ablation}. 
Table~\ref{tab:ablation} shows the contribution of each component of our framework.
By turning off the iterative learning module (IL), we see a consistently large performance drop for ITL, thus indicating IL's effectiveness. 
In addition, we ran two additional variants (\ie w/o MR-GAT and w/o MR prediction) to show the advantages of incorporating multi-relational room connections in both the encoder and the downstream task. In the first variant, we replace the multi-relation GAT in our encoder with vanilla GAT and update only spatial edges during training.
The second variant is done by predicting spatial, visual, and wall edges separately. Taking visual edge prediction as an example, the input graph is fed to the encoder and then one single decoder to predict the probability of the visual connectivity of two rooms. The same structure is applied to predict the other two edges as well.

\subsubsection{Effect of iteration} 
% \noindent\textbf{Model analysis.} 
Figure~\ref{fig:iter} shows the link prediction performance of ITL trained with different number of iterations.
The performance of the model reaches its peak after two iterations during training, and then gradually decreased.
We believe the drop is due to noisy messages beginning to permeate the entire graph and worsening the final prediction. The same phenomenon is also found in~\cite{xu2017scene,chen2020iterative}.
Note that in the proposed method, we do not share weights between blocks. $K$ iteration blocks are stacked together to give the final prediction. During our experiments, we found that not sharing weight gives us better performance (Table 8 in the supplementary material).
% (Table~\ref{tab:weight_sharing} in the supplementary material).

%% tab: feature set table
\begin{table}[t!]
    \caption{\textbf{Effect of different attribute sets.} Results are reported with ROC AUC. Set 1 includes basic room information (\eg room types). Set 2 includes distance-based features (\eg the length, width and area of rooms), Set 3 includes room shape descriptors (\eg chain code descriptors). Set 4 is image (\eg panorama) features of each room.}
    {\begin{center}
    \begin{adjustbox}{max width=1\linewidth}
        \begin{tabular}{cccc | ccc}
            \toprule
            set 1 & set 2 & set 3 & set 4 & Spatial & Wall & Door \tabularnewline
            \midrule
            \checkmark &&&& 78.75 & 63.76 & 91.04 \tabularnewline
            & \checkmark &&& 72.77 & 56.54 & 83.71 \tabularnewline
            && \checkmark && 72.01 & 56.44 & 84.00 \tabularnewline
            &&& \checkmark & 73.47 & 60.00 & 86.51 \tabularnewline
            \checkmark & \checkmark &&& 79.56 & \textbf{66.03} & 90.97 \tabularnewline
            \checkmark && \checkmark && 78.78 & 64.06 & 91.65 \tabularnewline
            \checkmark & \checkmark & \checkmark && 81.51 & 64.97 & 91.72
            \tabularnewline
            \checkmark & \checkmark & \checkmark & \checkmark & \textbf{81.64} & 65.67 & \textbf{91.73}
            \tabularnewline
            \bottomrule
        \end{tabular}
    \label{tab:comp_feat_set}
    \end{adjustbox}
    \end{center}
    }
\end{table}

\subsubsection{Contribution of various room attributes}
% \noindent\textbf{Contribution of various room attributes.}
The contribution and effect of different types of room attributes are shown in Table~\ref{tab:comp_feat_set}. It shows that each single attribute set can achieve fairly well performance, while the basic room information (\ie \textit{set 1}) contributes the most. It also shows that ITL does not require very specific room information to estimate the topology of floorplans and thus can be easily adopted by other indoor applications.
% Therefore, it can be easily applied many indoor applications. 
Still, with more detailed room information, ITL can achieve even higher performance.

\subsection{Limitations}\label{sec:limitation}
The input of our model is room-level attributes, which include basic room information, distance-based features, shape descriptors, and image features.
In this paper, all attributes are derived from human annotations. The attribute extraction process can be found in the supplementary material.
In practice, some of these room attributes might be difficult to obtain. 
One possible solution is to exclude the attributes that are too specific and hard to get. To this end, we study the effect of the different attributes (Table 5 of the main paper). We show that basic room information can achieve fairly well performance. However, with more detailed room information, ITL can achieve even higher performance.
Moreover, except human annotation, these attributes may also be extracted from state-of-the-art algorithms for scene classification (\eg~\cite{liu2019indoor, ismail2018understanding}) and room layout estimation (\eg~\cite{cabral2014piecewise,pintore2019automatic,yang2019dula}).

We also inherit other issues associated with ZInD~\cite{cruz2021zillow}. One issue is our architecture is trained on ZInD empty homes, which may not work as well with regular well-furnished homes. In addition, as discussed in \cite{cruz2021zillow}, there are ambiguities associated with large open spaces where, to quote \cite{cruz2021zillow}, ``semantic distinctions, such as `dining room', `living room' and `kitchen', are not always geometrically clear". In other words, room labels may not be consistent across different annotators, and may be another source of errors.

%% file: Supplementary.tex
% \documentclass[10pt,onecolumn,letterpaper]{article}

%%%%%%%%% PAPER TYPE  - PLEASE UPDATE FOR FINAL VERSION
% \usepackage[review]{cvpr}      % To produce the REVIEW version
% \usepackage{cvpr}              % To produce the CAMERA-READY version
% \usepackage[pagenumbers]{cvpr} % To force page numbers, e.g. for an arXiv version

% \usepackage{subcaption}
% \usepackage{graphicx}
% \usepackage{amsmath}
% \usepackage{amssymb}
% \usepackage{booktabs}

% \usepackage{tabularx, booktabs, ragged2e}
% \usepackage{adjustbox}
% \usepackage{multirow}
% \usepackage{lipsum}  
% \usepackage[ruled,vlined]{algorithm2e}
% \newcommand\mycommfont[1]{\footnotesize\ttfamily\textcolor{blue}{#1}}
% \SetCommentSty{mycommfont}

% \newcommand{\ie}{\textit{i}.\textit{e}., }
% \newcommand{\eg}{\textit{e}.\textit{g}., }
% \newcommand*{\etc}{etc.\@\xspace}%

% \newcommand{\singbing}[1]{{\textcolor{magenta}{[#1 -- Sing Bing]}}}
% \newcommand{\sbk}[1]{{\textcolor{magenta}{[#1 -- Sing Bing]}}}
% \newcommand{\Yu}[1]{{\textcolor{blue}{[#1 -- Yu]}}}
% \newcommand{\Will}[1]{{\textcolor{green}{[#1 -- Will]}}}
% \newcommand{\Ivo}[1]{{\textcolor{red}{[#1 -- Ivo]}}}
% \newcommand{\Naji}[1]{{\textcolor{cyan}{[#1 -- Naji]}}}

% %%
% %% end of the preamble, start of the body of the document source.
% \begin{document}

% \title{Supplementary Material}
% \section{Supplementary Material}

% \maketitle
% \addtocounter{table}{5}
% \addtocounter{figure}{7}

We provide details of the dataset (Section~\ref{sec:dataset}), implementation (Section~\ref{sec:implementation}) and further demonstrate the effectiveness of the proposed method with additional floorplan generation results (Section~\ref{sec:additional_results}).
We visualize the graph topology of each iteration in Section~\ref{sec:graphVis}.
In Section~\ref{sec:sensitivity}, we show results of sensitivity analysis of hyper-parameters, and compare the effect of raw and learnt features for more insights on our proposed model.  Section~\ref{sec:longskip} shows the benefit of having long skip connections and weight sharing in our method.
% Finally, limitations of the work are discussed in Section~\ref{sec:limitation}.

\section{Dataset}\label{sec:dataset}
In this section, we provide more details of the dataset (e.g., node, edge, label distributions), and the procedure of graph construction.
\subsection{Data statistics}
\begin{figure*}[t]
    \centering
    \includegraphics[trim=0in 0in 0in 0in,clip,width=1\linewidth]{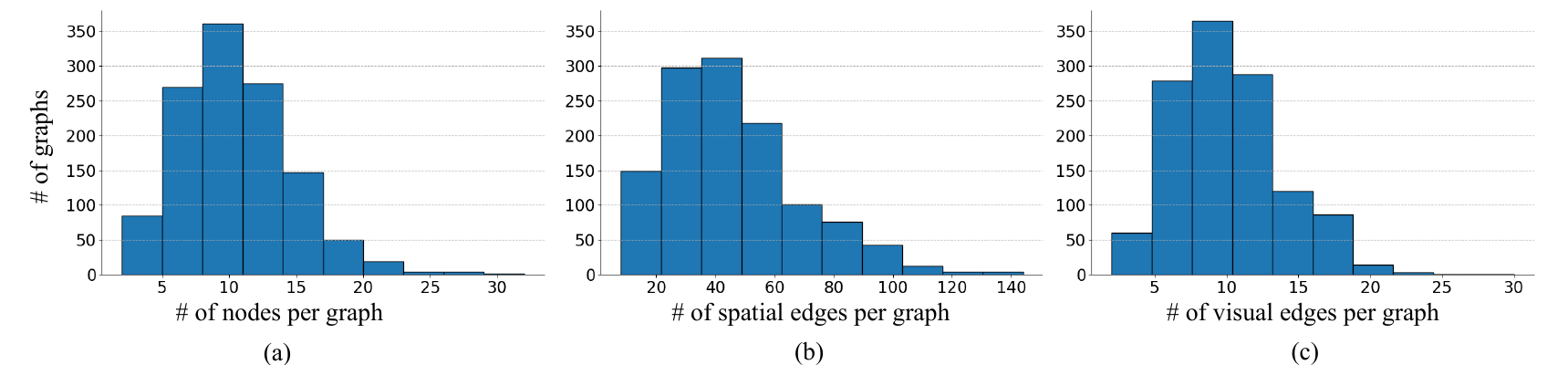}
    \caption{Node and edge distribution}
    \label{fig:node_edge_dist}
\end{figure*}

\begin{figure*}[t]
    \centering
    \includegraphics[trim=0in 0in 0in 0in,clip,width=0.95\linewidth]{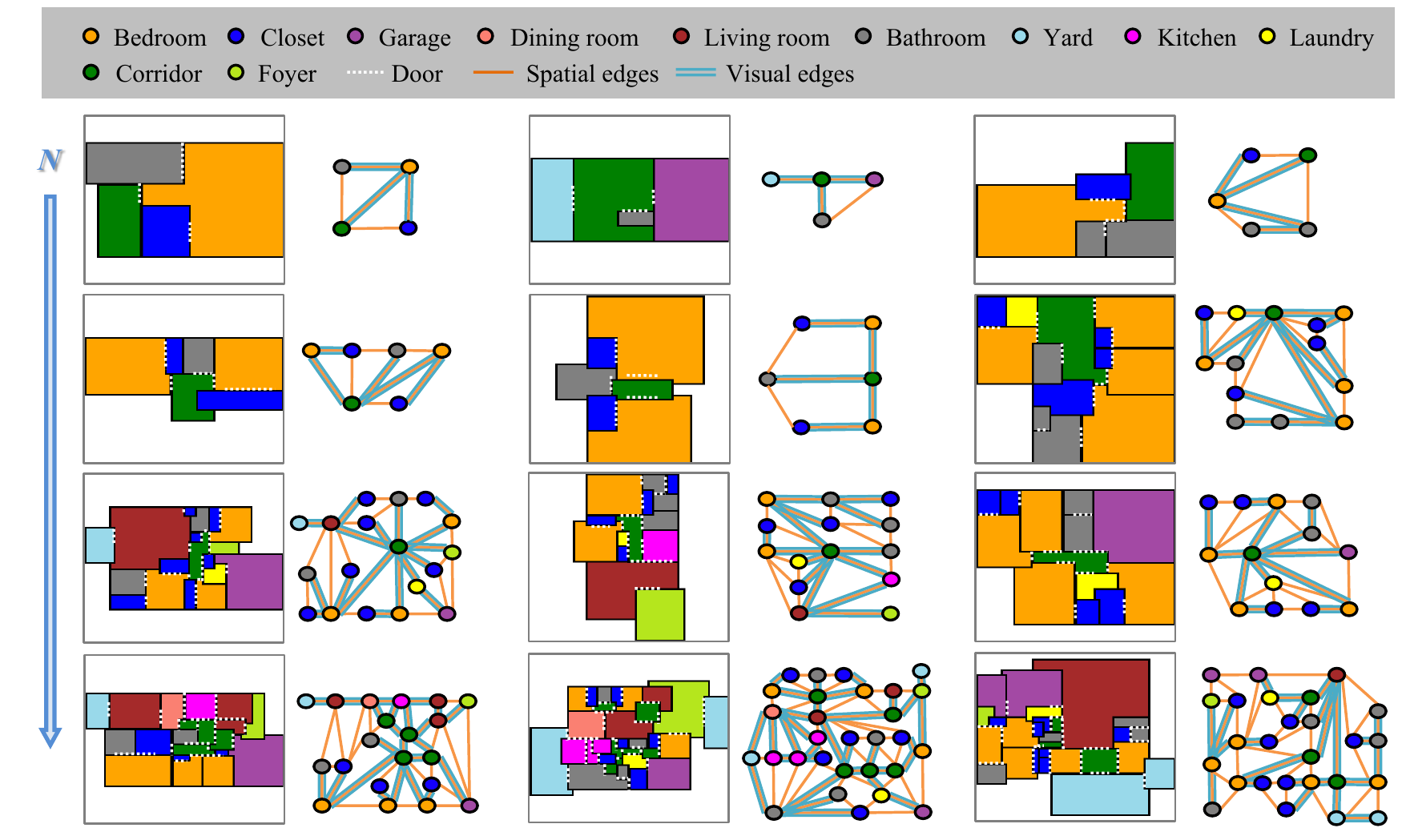}
    \caption{Examples of floorplans in dataset and corresponding constructed graphs. $N$ refers to the number of nodes. The floorplans are arranged in ascending order of complexity.}
    \label{fig:dataSamples}
\end{figure*}

The Zillow Indoor Dataset (ZInD)~\cite{cruz2021zillow} contains 2,737 floorplans, from which we retrieved 1,217 and constructed graphs containing nodes and multi-relational edges. 
We show node and edge distributions in Fig.~\ref{fig:node_edge_dist}. The dataset includes different floorplans with a wide range of nodes and edges. We can see that most of the graphs have around 10 nodes; however, the most complicated floorplans contain around 30 rooms, 140 spatial edges, and 30 visual edges. We can also see that, in general, the number of spatial edges is 4-5 times more than that of visual edges. Some floorplan examples of the dataset are shown in Fig.~\ref{fig:dataSamples}.

%% figure: room type distribution
\begin{figure*}[h!]
    \centering
    \includegraphics[trim=0in 0in 0in 0in,clip,width=0.65\linewidth]{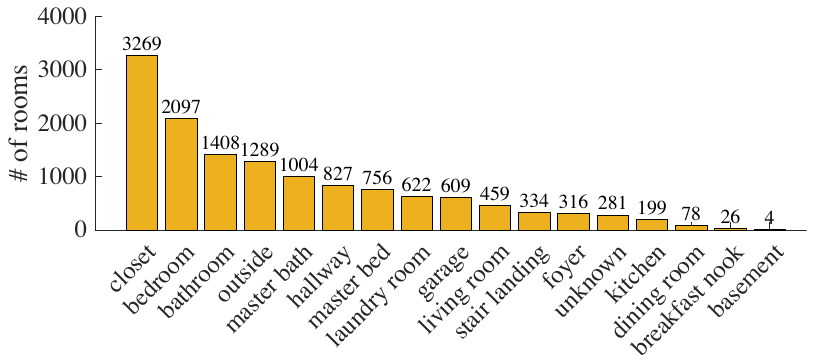}
    \caption{Frequency distribution of room types.}
    \label{fig:room_type}
\end{figure*}

Fig.~\ref{fig:room_type} shows the frequency distribution of the room types. The dataset includes 17 different types of rooms. The top three most frequent labels are closet, bedroom, and bathroom. The top three least frequent labels are basement, breakfast nook, and dining room. In many cases, breakfast nooks are labeled as kitchens, and dining rooms are merged into living rooms, thus reducing the frequency of these two labels.

%% figure: node, edge distribution

\begin{figure*}[t]
  \centering
  \begin{subfigure}{0.32\linewidth}
    \centering
    \caption{Example image of one bedroom.}
    \includegraphics[trim=0in 0in 0in 0in,clip,width=1\linewidth]{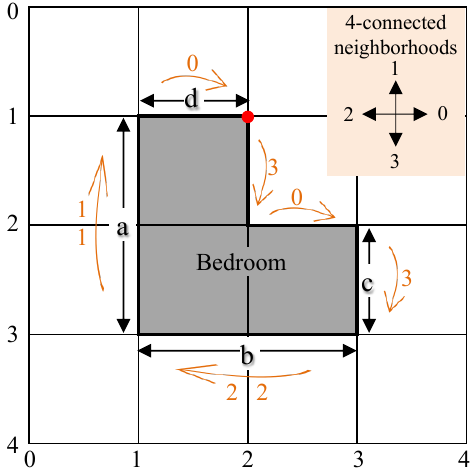}
    
    \label{fig:attr_details}
  \end{subfigure}
  \hfill
  \begin{subfigure}{0.62\linewidth}
    %% feat
    \centering
    \caption{Detailed description of four attribute sets. }
    \begin{adjustbox}{max width=\linewidth}
    \begin{tabular}{r|cc}
    \toprule
     & Name & Value \\
    \midrule
    \multirow{4}{*}{Basic room information} & {room type} & "bedroom"\\ 
    \multirow{4}{*}{} & \# of doors & 1 \\
    \multirow{4}{*}{} & \# of windows & 2 \\
    \multirow{4}{*}{} & \# of openings & 0 \\
    \midrule
    \multirow{4}{*}{Distance-based features} & {room  perimeter} & $2a+2b$\\ 
    \multirow{4}{*}{} & maximum length \& width & $(a,b)$ \\
    \multirow{4}{*}{} & room area & $a*b$ \\
    \multirow{4}{*}{} & $\frac{\text{room area}}{\text{its bounding box area}}$ & $\frac{a*b-(a-c)*(b-d)}{a*b}$ \\
    \midrule
    \multirow{1}{*}{Shape descriptors} & chain code & 03033133-1-1-1...-1 \\
    \midrule
    \multirow{1}{*}{Image features} & panoramas features & $f(I_{pano})$ \\
    \bottomrule
    \end{tabular}\label{tbl:attr_details} 
    \end{adjustbox}

  \end{subfigure}
  \caption{Example of the attributes extracted from a room.}
  \label{fig:short}
\end{figure*}

\subsection{Attribute extraction}\label{sec:attr_extraction}
Four types of attributes, including basic room information, distance-based features, shape descriptors, and image features, are extracted for each room. We concatenate features from different attribute sets and then feed them into the network. Detailed description of these attributes are listed in Table~\ref{tbl:attr_details}.

We use the Chain code~\cite{freeman1961encoding} to represent the shape of rooms. 
To generate a scale and rotation invariant shape descriptor, we first adopt the rule of 4-connected neighbourhood to traverse the boundary of the room, and record the direction of each step. 
As shown in Fig~\ref{fig:attr_details}, we first choose a start position (\ie (2,1)) and develop the chain code from it:
$$
c = (3,0,3,2,2,1,1,0)
$$

To make the chain code invariant to rotation, we compute the difference between elements in the code. So, each element can be computed as:

\begin{equation}
  R(c_i) =
    \begin{cases}
      mod\{(c_{i+1} - c_i), 4\}, & (0\leq i \leq N-1)\\
      mod\{(c_{0} - c_i), 4\}, & (i = N-1)\\
    \end{cases}       
\end{equation}

Then, we make the chain code invariant to start position by applying the minimum circular shift. Specifically, we rotating the elements in the chain code until it gives the smallest integer. From Fig~\ref{fig:attr_details}, we can get
$R(c) = (1,3,3,0,3,0,3,3)$ and $S(R(c)) = (0,3,0,3,3,1,3,3)$. Finally, we pad the chain code to a fixed length of 100 with -1.

Image features are extracted from a scene classification model using the NASNet model~\footnote{https://www.tensorflow.org/api\_docs/python/tf/keras/applications/nas net/NASNetMobile}~\cite{zoph2018learning} followed by an average pooling layer and a fully connected layer for prediction. The features are extracted from the output of the last pooling layer and have a size of 1056. The classification model is pre-trained on ImageNet and then retrained on ZInD~\cite{cruz2021zillow}.

%% figure: data samples

%% figure: spaVSvis
\begin{figure*}[t]
    \centering
    \includegraphics[trim=0in 1.5in 0in 0in,clip,width=0.55\linewidth]{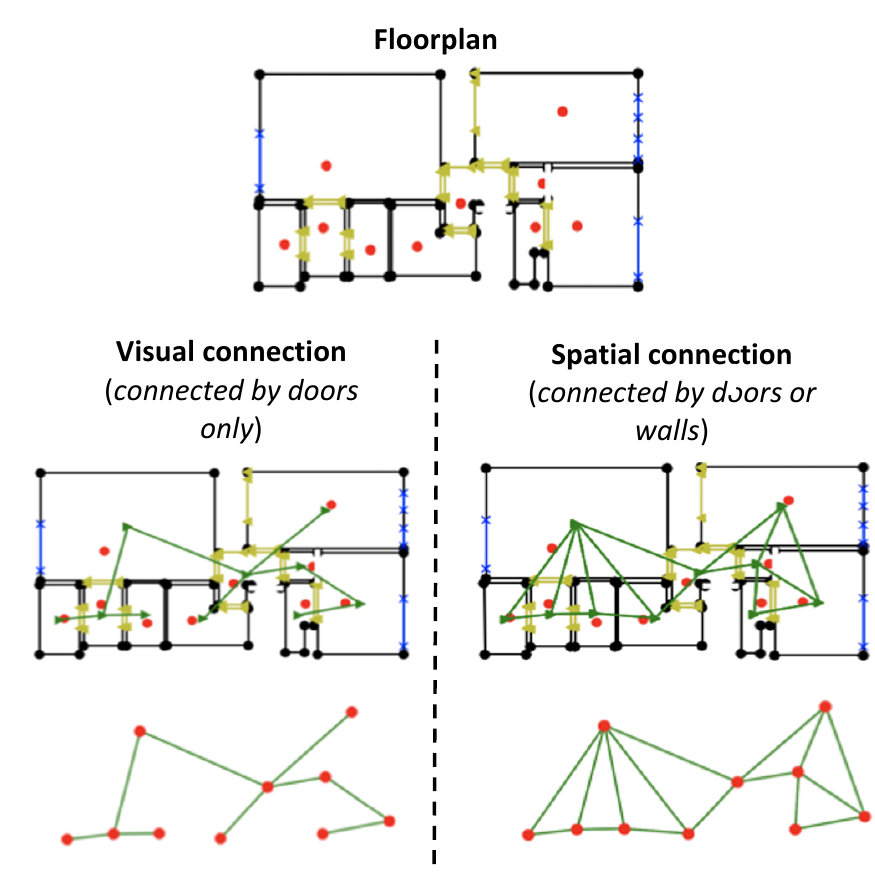}
    \caption{Graph construction. Black, yellow, and blue lines in the floorplan refers to walls, doors, and windows, respectively. From the floorplan, we extract spatial and visual connections to build a multi-relational graph. The attribute of rooms (\eg number of doors, windows, the area of the room) are also computed from the floorplan.}
    \label{fig:spaVSvis}
\end{figure*}

%% figure: HouseGAN results
\begin{figure*}[t]
    \centering
    \includegraphics[trim=0in 0in 0in 0in,clip,width=1\linewidth]{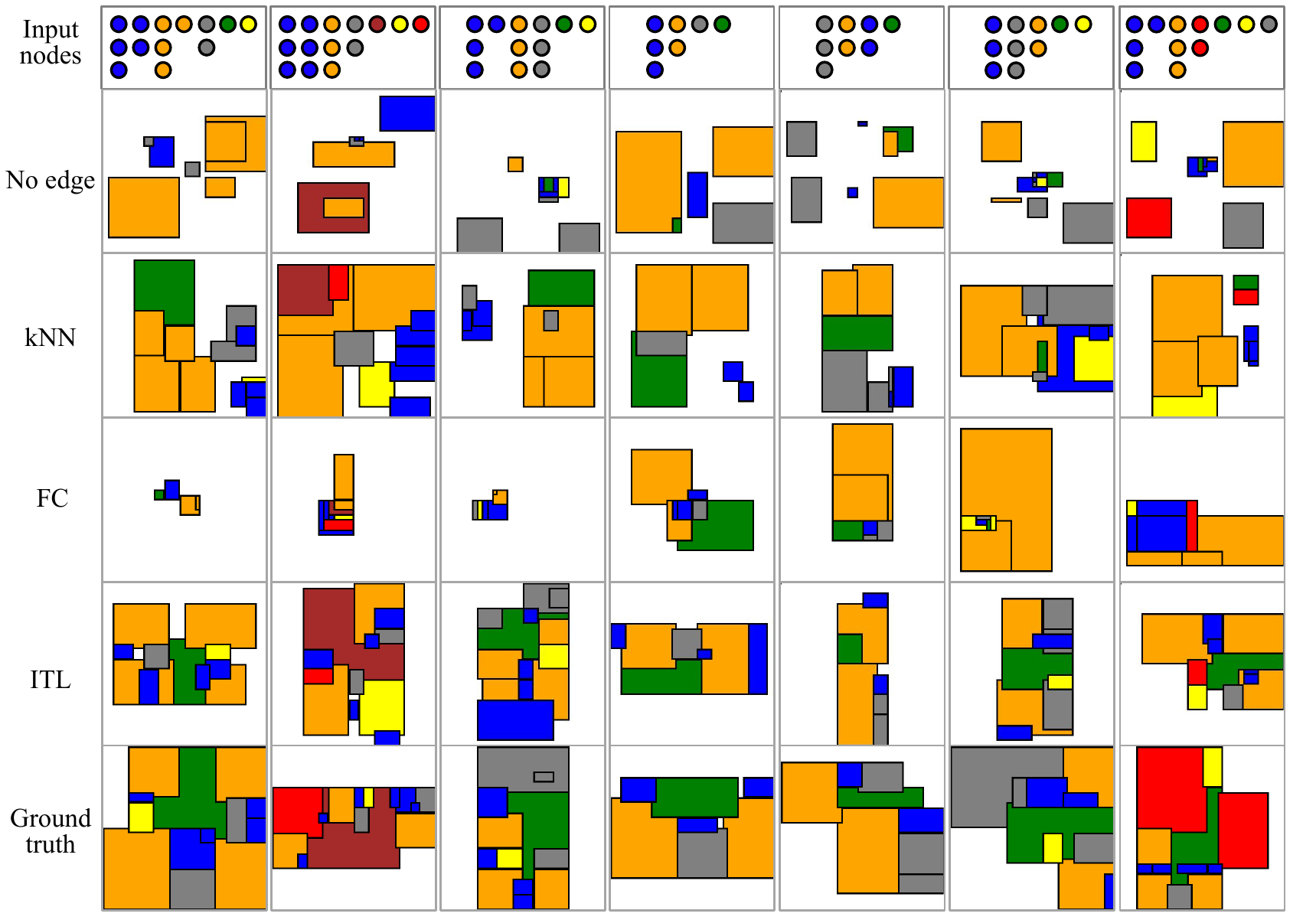}
    \caption{Comparison of the usage of different graph structures (\ie no edge given, kNN graphs, fully connected (FC) graphs, topology estimated by ITL) in the floorplan generation task.}
    \label{fig:housegan_supple}
\end{figure*}

\subsection{Graph construction}
The floor plan can easily and naturally be represented as a graph. Each node represents a room, and edges represent the connectivity of two rooms. We construct graph edges by defining two types of edges, which are spatial edges and visual edges (Fig.~\ref{fig:spaVSvis}). More samples of constructed graph are shown in Fig.~\ref{fig:dataSamples}.

\section{Implementation details}\label{sec:implementation}
Our model has 2 iterative encoder blocks ($K=2$). In each block, the graph attention layer applied on each subgraph has the dimension of 32.
The three fully connected layers in decoder has the dimension of 64, 16, and 1, respectively.
Adam optimizer is adopted to train the model, with a learning rate of $1\times 10^{-4}$ that dropped 0.5 at $60^{th}$ and $120^{th}$ epochs.
The model is trained for 300 epochs with early stopping.
Implementation was done using PyTorch and trained with a Nvidia TITAN-XP GPU.

\section{Floorplan generation}
\label{sec:additional_results}
We show more qualitative comparison of different graph structures (\ie no edge given, kNN graphs, fully connected (FC) graphs, topology estimated by ITL) usage for floorplan generation. We show that ITL can generate effective and practical floorplan topological information, which leads to more realistic floorplan generation using HouseGAN~\cite{nauata2020house}.
We can see that, without topological information (\ie no edge provided), most of rooms in generated floorplan are separated.
The kNN graph fails to provide compelling topology for HouseGAN, since it predicts connections based on the similarity of room attributes. Rooms generated with kNN graphs tend to be clustered by the room type (\eg bedrooms are connected to bedrooms, closets are connected to closets). Furthermore, the fully connected graph will cause spatial overlap of rooms.

%% figure: iter_vis
\begin{figure*}[t]
    \centering
    \includegraphics[trim=0in 0in 0in 0in,clip,width=0.8\linewidth]{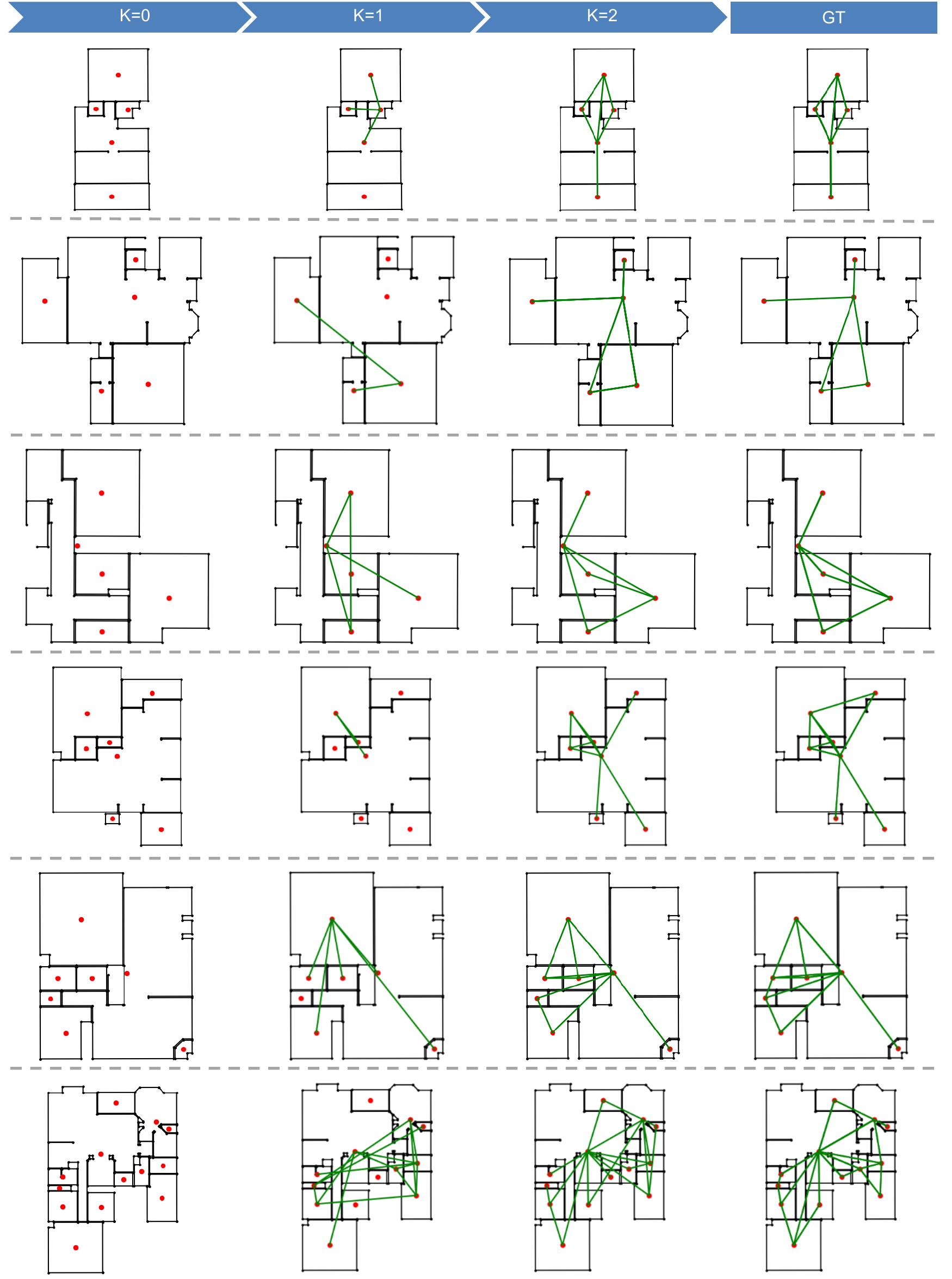}
    \caption{Examples of floorplans in dataset and corresponding constructed graphs. $N$ refers to the number of nodes. The floorplans are arranged in ascending order of complexity.}
    \label{fig:iter_vis}
\end{figure*}

\section{Graph visualization}\label{sec:graphVis}
We visualize the learned graph topology in each iteration. As shown in Fig.~\ref{fig:iter_vis}, the initial graph topology is completely missing (\ie $k=0$). After the first iteration (\ie $k=1$), the model predicts a subset of the edges based only on node embeddings. 
Then the second iteration (\ie $k=2$) will further refine the results and complement the predictions in the first iteration.

\section{Sensitivity analysis}
\label{sec:sensitivity}

We predict multiple types of edges (\ie spatial and visual edges) in this work. We split spatial edges into two exclusive sets: door (\ie visual) connectivity and wall connectivity, and then jointly predict all three types of edges by combining the loss function of each task. The loss function $\mathcal{L}$ of the proposed model can be expressed as:
\begin{equation}
\mathcal{L} = \mathcal{L}_{vis} + \lambda_1 \mathcal{L}_{wall}  + \lambda_2 \mathcal{L}_{spa} ,
\end{equation}
where $\mathcal{L}_{vis}$, $\mathcal{L}_{wall}$, and $\mathcal{L}_{spa}$ denote the binary cross-entropy loss for visual (door), wall, and spatial connectivity, respectively. $\lambda_1$ and $\lambda_2$ are two non-negative hyper-parameters in our model. We used $\lambda_1=\lambda_2=1$ in our paper.

To give more insight of the model, we conduct sensitivity analysis of these two hyper-parameters. Results are summarized in Table~\ref{tab:sensitive_analysis}. We can see that the overall performance of the model is stable. When $\lambda_2$ remains the same, a larger $\lambda_1$ will improve the prediction performance for all types of edges. The improvement slows down if $\lambda_1$ is greater than 1. 
When $\lambda_1$ remains the same, a larger $\lambda_2$ will lead to higher performance of the model. If $\lambda_2$ is greater than 1, although the performance of door prediction is slightly improved, the performance of wall prediction will be greatly reduced, resulting in a decrease in overall performance.

%% tab: sensitive analysis
\begin{table}[t]
\caption{Sensitivity analysis, with ROC AUC being the metric. In our work, we use $\lambda_1=1$, $\lambda_2=1$ (with the results in bold).}
\label{tab:sensitive_analysis}
\begin{center}
\begin{adjustbox}{max width=0.8\linewidth}
\begin{tabular}{c | c c c}
\toprule
  & Spatial & Wall & Door\\
\hline
$\lambda_1=0.25$, $\lambda_2=1$ & 79.16 & 61.60 & 89.62 \\
$\lambda_1=0.5$, $\lambda_2=1$ & 80.91 & 64.20 & 90.11 \\
$\lambda_1=1$, $\lambda_2=1$ & \textbf{81.64} & \textbf{65.67} & \textbf{91.73} \\
$\lambda_1=2$, $\lambda_2=1$ & 81.00 & 65.70 & 91.63 \\
\hline
$\lambda_1=1$, $\lambda_2=0.25$ & 79.05 & 64.22 & 90.89 \\
$\lambda_1=1$, $\lambda_2=0.5$ & 80.38 & 64.75 & 90.63 \\
$\lambda_1=1$, $\lambda_2=1$ & \textbf{81.64} & \textbf{65.67} & \textbf{91.73} \\
$\lambda_1=1$, $\lambda_2=2$ & 80.21 & 62.66 & 92.54 \\
\bottomrule
\end{tabular}
\end{adjustbox}
\end{center}
\end{table}

\section{Effect of long skip connection and weight sharing}
% \section{Effect of long skip connection}
\label{sec:longskip}
Here, we demonstrate the effect of long skip connection in the proposed model. Table~\ref{tab:longSkipConnection} shows the results with and without long skip connection. We set the initial room attributes (\ie raw features) as the baseline and directly feed them into the decoder to predict multiple relationships between rooms. Then we remove the long skip connection in the proposed model, and predict edges using only learned features. Finally, we report results of using combined features (\ie with long skip connection). The comparison of the raw, learned, and combined features are summarized in Table~\ref{tab:longSkipConnection}.

Note that in the proposed method, we do not share weights between blocks. $K$ iteration blocks are stacked together to give the final prediction. During our experiments, we found that not sharing weight gives us better performance (Table~\ref{tab:longSkipConnection}).

%% tab: feat_comparison
\begin{table}[h]
\caption{Comparison of raw and learnt features, with accuracy metric being ROC AUC.}\label{tab:longSkipConnection}
\begin{center}
\begin{adjustbox}{max width=1\linewidth}
\begin{tabular}{c | c | c c c}
\toprule
  & features & Spatial & Wall & Door\\
\hline
baseline & raw & 70.75 & 54.91 & 80.85 \\
w/o long skip connection& learned & 75.64 & 60.66 & 90.37 \\
ITL& combined & \textbf{81.64} & \textbf{65.67} & \textbf{91.73} \\
\bottomrule
\end{tabular}
\end{adjustbox}
\end{center}
\end{table}

%% tab: weight_sharing 
\begin{table}[t!]
\caption{Effect of weight sharing, with ROC AUC being the metric.}
\label{tab:weight_sharing}
\begin{center}
\begin{adjustbox}{max width=0.8\linewidth}
\begin{tabular}{c | c c c}
\toprule
 & Spatial & Wall & Door\\
\hline
\multirow{1}{*}{shared weights} &  72.38 & 60.99 & 87.13 \\
\multirow{1}{*}{different weights} & \textbf{81.64} & \textbf{65.67} & \textbf{91.73} \\
\bottomrule
\end{tabular}
\end{adjustbox}
\end{center}
\end{table}

% %%%%%%%%% REFERENCES
% {\small
% \bibliographystyle{ieee_fullname}
% \bibliography{egbib}
% }

% \end{document}